# Developing a Novel Image Marker to Predict the Clinical Outcome of Neoadjuvant Chemotherapy (NACT) for Ovarian Cancer Patients

Ke Zhang[a,b*], Neman Abdoli[b*], Patrik Gilley[b], Youkabed Sadri[b], Xuxin Chen[b], Theresa C. Thai[c], Lauren Dockery[d], Kathleen Moore[d], Robert S. Mannel[d], Yuchen Qiu[a,b,1]

[a] Stephenson School of Biomedical Engineering, University of Oklahoma, Norman, OK, USA 73019
[b] School of Electrical and Computer Engineering, University of Oklahoma, Norman, OK, USA 73019
[c] Department of Radiology, University of Oklahoma Health Sciences Center, Oklahoma City, OK, USA 73104
[d] Department of Obstetrics and Gynecology, University of Oklahoma Health Sciences Center, Oklahoma City, OK, USA 73104

## ABSTRACT

**Objective:** Neoadjuvant chemotherapy (NACT) is one kind of treatment for advanced stage ovarian cancer patients. However, due to the nature of tumor heterogeneity, the clinical outcome to NACT vary significantly among different subgroups. The patients with partial responses to NACT may lead to suboptimal debulking surgery, which will result in adverse prognosis. To address this clinical challenge, the purpose of this study is to develop a novel image marker to achieve high accuracy prognosis prediction of the NACT at an early stage.

**Methods:** For this purpose, we first computed a total of 1373 radiomics features to quantify the tumor characteristics, which can be grouped into three categories: geometric, intensity, and texture features. Second, all these features were optimized by principal component analysis algorithm to generate a compact and informative feature cluster. This cluster was used as input for developing and optimizing support vector machine (SVM) based classifiers, which indicated the likelihood of the patient receiving suboptimal cytoreduction after the NACT treatment. Two different kernels for SVM algorithm were explored and compared. To validate this scheme, a total of 42 ovarian cancer patients were retrospectively collected. A nested leave-one-out cross-validation framework was adopted for model performance assessment.

**Results:** The results demonstrated that the model with a Gaussian radial basis function kernel SVM yielded an AUC (area under the ROC [receiver characteristic operation] curve) of 0.806 ± 0.078. Meanwhile, this model achieved overall accuracy (ACC) of 83.3%, positive predictive value (PPV) of 81.8%, and negative predictive value (NPV) of 83.9%.

**Conclusion:** This study provides meaningful information for the development of radiomics based image markers in NACT treatment outcome prediction.

---

[1] Corresponding author: qiuyuchen@ou.edu   *Both Ke Zhang and Neman Abdoli are considered first authors.

# 1. INTRODUCTION

Being the most lethal gynecological malignancy, ovarian carcinomas are anticipated to result in approximately 19,710 new cases and 13,270 deaths within the US during 2023 [1]. Among all the diagnosed patients, approximately 90% are categorized as epithelial carcinoma, which can be further divided into four subtypes, namely, serous, endometrioid, clear cell and mucinous carcinomas [2]. Meanwhile, the remaining 10% patients are diagnosed with non-epithelial ovarian cancers, which includes germ-cell or sex cord stromal tumors [2]. Given that there are not effective early stage screening approaches, most ovarian cancer patients are diagnosed at an advanced stage (III or IV) [3, 4]. Currently, the standard of care is to first perform primary debulking surgery (PDS) to remove the visible tumors. Then adjuvant chemotherapy (ACT) is followed up to control the residual invisible tumor cells. Recently, one alternative has emerged, suggesting the initiation of neoadjuvant chemotherapy (NACT) [5] before cytoreductive surgery (i.e. interval debulking surgery [IDS]) and additional follow-up chemotherapy. Some investigations have indicated its advantages for certain groups of patients who are not well-suited for PDS [6, 7].

However, due to the tumor heterogeneity, the tumor responses to NACT therapy vary significantly across different patient subcategories [8]. Therefore, the well responsive patients can receive optimal debulking in IDS with either no visible residual tumors or residual tumors measuring less than 1 cm in diameter, which will obtain the best overall survival after the following treatment [9-11]. Nevertheless, the remaining patients experiencing partial response may have to undergo suboptimal debulking treatment, leaving residual tumors larger than 1 cm in diameter. Therefore, to achieve an optimal treatment effect, it is of crucial importance to develop a prognostic model which is able to identify these patients who have to receive suboptimal debulking surgery after NACT therapy. To date, different approaches have been explored to predict the outcome of ovarian cancer treatments, which are based on FIGO (International Federation of Gynecology and Obstetrics) staging system [12], demographic information [13], radiology findings [14, 15], CA-125 (cancer antigen 125) [16, 17], HE4 (Human Epididymis Protein 4) [18], BRCA 1/2 gene mutations [19], and so forth. However, due to the limited reliability and robustness of these models, they are not yet widely recognized [14, 20] in clinical practice.

Among all those technologies used in the prognostic assessment, the computed tomography (CT) scan [21] has its unique superiority over the others due to its advantages of wide availability and low operating cost [22, 23]. Meanwhile, radiomics [24-26], a recent emerging approach, can extract and quantify the tumor characteristics from CT images, aiming to uncover the underlying biological mechanism associated with patient's diagnosis or prognosis. However, few studies have been conducted on using these novel techniques for clinical outcome prediction in NACT treatment for ovarian cancer patients. In this investigation, our objective is to develop a radiomics based CT image marker to accurately identify the patients who may not be able to receive optimal cytoreduction after NACT treatment. The major contributions of this study are summarized as follows:

- As far as we know, this study is one of the pioneering investigations which developed a CT image based clinical marker to predict outcomes of IDS for ovarian cancer patients.
- We discovered a unique radiomics feature signature that has a strong association with IDS. This feature signature was identified from a comprehensive set of 1373 radiomics features which can be categorized into 3 groups, namely, shape, density, and texture features.
- We developed a clinical marker based on the feature signature. The effectiveness of this marker was evaluated by a dataset containing a total of 42 patients. The model achieved an overall accuracy of 83.3%, and an area under the ROC (receiver characteristic operation) curve of $0.806 \pm 0.078$.

The details are presented in the following sections.

## 2. MATERIALS AND METHODS

### 2.1 Database

This study was approved by the Institutional Review Board (IRB 13649) at the University of Oklahoma. The patients' consents were not needed as all the image data were collected retrospectively. This dataset contains a total of 42 advanced stage ovarian cancer patients received NACT regimens. Among them, 28 patients underwent optimal debulking in the following IDS, while the remaining patients did not. All the selected patients were diagnosed with recurrent epithelial ovarian cancer (EOC). Their treatments were performed at the University of Oklahoma Health Sciences Center. The patients' histology types include endometrial, serous, clear cell, and mucinous. For each patient, we collected the pre-therapy CT images following the standard acquisition protocols. Specifically, the image obtainment was conducted on a GE LightSpeed VCT 64-detector or GE Discovery 600 16-detector CT machines. X-ray tube current was set to be from 100 to 600 mA to fit the various body size, with a tube power of 120 kVp. The examination was conducted by the following procedures: The contrast agent (Isovue 370, 100 cc) was first injected to the patient intravenously at a rate of 2–3 cc per second. The initial scan started 60 seconds after injection began, while the second scan started 5 minutes after the completion of injection. CT images were reconstructed to 1.25 mm in axial axis, and 2.5 mm in sagittal and coronal directions.

### 2.2 Image Feature Computation

Prior to the image feature computation, we performed tumor segmentation on each case to generate the 3D volume of interest (VOI). An experienced radiologist identified each tumor on the acquired CT images and attached annotation to the image showing the largest tumor area. With the annotated reference images, all the slices containing metastatic tumors can be determined accurately and processed by a previously developed segmentation algorithm [27], which combines a multilayer topographic region growth algorithm with adaptive thresholds and a dynamic edge tracking method. Due to the heterogeneity of metastatic tumors, the automated segmentation results may not be adequately accurate for the following process. Therefore, the automatic segmentation results were visually evaluated by experienced researchers and the tumor contours were manually adjusted as needed.

Next, a large amount of radiomics features were computed from the segmented tumors in the original CT images, using the open-source platform Pyradiomics v3.0.1 [28]. Three categories of features were computed: shape, first order, and texture features. The shape features characterize the 3D size and geometry of the lesion from diverse perspectives, such as voxel volume and elongation. First order features depict the intensity distribution of the voxels inside the segmented VOI. The texture features include gray level co-occurrence matrix (GLCM) features [29], gray level dependence matrix (GLDM) features [30], gray level run length matrix (GLRLM) features [31], gray level size zone matrix (GLSZM) features [32], and neighboring gray tone difference matrix (NGTDM) features [33]. These features enable a quantitative depiction of the tumor textures by assessing the spatial distribution and variation of voxel values. For instance, GLDM features are generated from a matrix that describes the appearance of qualified neighboring voxels surrounding the central voxel to measure the dependency relationships across intensity levels. Furthermore, we also obtained extra first order and texture features from several filtered image types: exponential, gradient, local binary pattern 3D (LBP3D), logarithm, square, square root, and wavelet filters. In summary, a total of 1373 features extracted from the smallest tumor in each case were utilized for training the model, as illustrated in table 1.

Table 1. The number of features of different image types and feature types used in this study.

| Image Type | Feature Type | | | | | | | Total |
|---|---|---|---|---|---|---|---|---|
| | Shape | First order | GLCM | GLDM | GLRLM | GLSZM | NGTDM | |
| Original | 14 | 18 | 24 | 14 | 16 | 16 | 5 | 107 |
| Exponential | 0 | 16 | 0 | 7 | 8 | 3 | 0 | 34 |
| Gradient | 0 | 18 | 24 | 14 | 16 | 16 | 5 | 93 |
| LBP3D | 0 | 45 | 21 | 14 | 16 | 16 | 5 | 117 |
| Logarithm | 0 | 18 | 24 | 14 | 16 | 16 | 5 | 93 |
| Square | 0 | 17 | 24 | 14 | 16 | 16 | 5 | 92 |
| Square root | 0 | 18 | 24 | 14 | 16 | 16 | 5 | 93 |
| Wavelet | 0 | 144 | 192 | 112 | 128 | 128 | 40 | **744** |
| Total | 14 | 294 | **333** | 203 | 232 | 227 | 70 | 1373 |

## 2.3 Develop a Machine Learning based Model to Predict the Clinical Outcome

Using the established feature pool, we developed a classification model to predict whether the disease can be optimally debulked in the IDS (Figure 1). At the onset of the model, all the features were first scaled by the formula: $\mathbf{x}' = (\mathbf{x} - \mu)/\sigma$, where $\mathbf{x}$ is a vector containing original values of a feature, $\mu$ and $\sigma$ are the corresponding mean and standard deviation calculated from training set, respectively. Given the imbalanced nature of the dataset, SMOTE (synthetic minority over-sampling technique) was employed to augment data in the minority class [34], which is based on the imbalanced-learn library v0.9.0 [35]. In this method, it first selects a sample $\mathbf{x}'_r$ from the randomly shuffled minority class. Using $\mathbf{x}'_r$ as a reference point, the algorithm locates its $k$ nearest neighbors among minority group. A new sample $\mathbf{x}'_s$ will then be synthesized by the formula: $\mathbf{x}'_s = \mathbf{x}'_r + \beta(\mathbf{x}'_i - \mathbf{x}'_r)$, where $\mathbf{x}'_i$ is one of the $k$ nearest neighbors and $\beta$ is randomly selected from the interval (0, 1). This synthetic process will persist until the minority dataset is augmented to the balanced level.

Next, feature dimension reduction was conducted using principal component analysis (PCA) [36] on the matrix $\mathbf{X}'$ containing features of all training instances. As a popular method, PCA can effectively reduce the dimension while retaining variance as much as possible [37], especially when the dataset is relatively small [38]. Accordingly, PCA generates the principal components (PCs) based on the co-variance matrix $\mathbf{M} = \mathbf{X}'^T\mathbf{X}'$. Each PC is an eigenvector of the matrix M representing a direction in the redefined feature space, while the variance explained by this PC is represented by its corresponding eigenvalue. Based on the spectral theorem [39], these variances quantify the importance of each PC to the matrix. Since the variance decreases significantly, it becomes feasible to use only a small portion of the most important PCs to approximate the original feature vectors with a satisfactory accuracy [36]. We used PCA to reduce the original feature set to less than 10 features prior to classification.

After that, a support vector machine (SVM) classifier [40] was implemented using scikit-learn v1.0.2 library [41], which aims to differentiate between sub-optimally debulked and optimally debulked cases within the dataset. As a robust classification algorithm, SVM serves well in other related studies of prognostic modeling of NACT [42-44] and has demonstrated its effectiveness in small size dataset based classification tasks [45]. In binary classification tasks, rather than just finding one hyperplane that simply separates the two classes, SVM seeks an optimal hyperplane that has the largest margin between the two classes, relying on the samples closest to the hyperplane (i.e., support vectors). The SVM classifier with linear kernel (Linear-SVM) or Gaussian radial basis function kernel (RBF-SVM) was adopted in the classification task. The regularization parameters C for both kernels were searched among a logarithmic array: $10^{-5}$, $10^{-4}$, …, $10^2$. For RBF kernel specifically, its kernel coefficient gamma was chosen from another logarithmic array: $10^{-3}$, $10^{-2}$, …, 10. The pipeline of this machine learning model is depicted in Fig. 1a.

Finally, we employed a nested leave-one-out cross-validation (LOOCV) [46, 47] to evaluate the classification performance of the SVM-based clinical outcome prediction models (Figure 1b). LOOCV is a unique type of cross-validation technique, in which the testing set only contains one case and all the remaining cases are used to train the model. This approach can maximally mitigate the influence of random partitioning of dataset, ensuring an almost unbiased evaluation result [48]. However, since hyperparameter tuning was also implemented within the LOOCV process, it incorporates testing data into model selection, which may result in overestimating the model's generalization performance. Thus, we adopted the nested LOOCV with 2 levels: The outer LOOCV is sorely responsible for the model evaluation; the inner LOOCV is embedded within the outer LOOCV to fine-tune the hyperparameters of each module in the model, including SMOTE, PCA, and SVM. This nested structure effectively decouples the testing data from training data, leading to a significant decrease in evaluation bias, and bringing error estimation close to the true level. The model performance was assessed by the receiver operating characteristic (ROC) curve [49], and various metrics such as the area under curve (AUC) [50] value, overall accuracy (ACC), positive predictive value (PPV), and negative predictive value (NPV). Overall, the model employed in this preliminary study has limited complexity given the small size of dataset, which aims to minimize the risk of learning from noise rather than the actual patterns. We utilized PCA for dimension reduction and SVM for prognosis classification, recognizing their simplicity and robustness. Similar methodologies have been adopted successfully in prior radiomics based classification studies [51, 52].

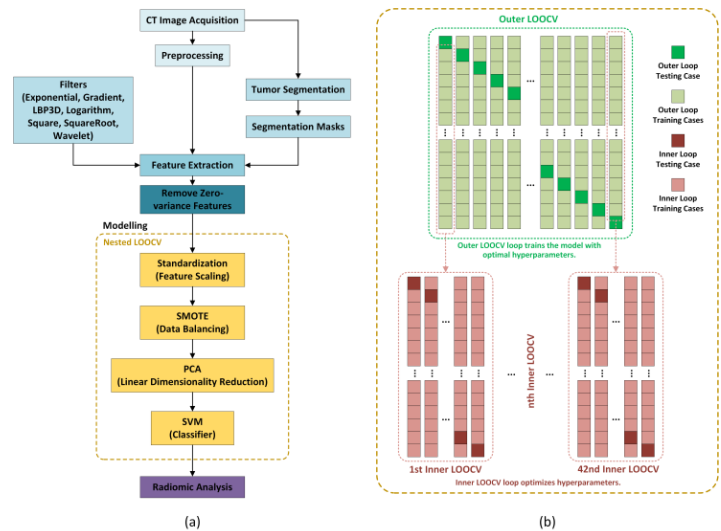

Fig. 1. Pipeline of the radiomic analysis in this study.

## 3. RESULTS

Figure 2 demonstrates one sample case of our tumor segmentation results as well as 6 of the extracted 1373 features from this tumor. All the CT images containing this tumor are displayed from left to right in the first row, and the outline of the ROI tightly wrapping the tumor tissue in each image is identified and marked by red color. The tumor in each CT image is isolated and illustrated in the second row which contains all the information that is going to be extracted for this case. The 3D shape of this tumor is shown in the third row, accompanied by 6 of its logarithm and gradient features.

The hierarchically clustered heatmap [53] of Pearson's correlation coefficients for the extracted 1373 radiomic features is demonstrated in Figure 3, with the values ranging from -1 to 1. The negative value is represented using a brown color, while the positive value is represented by a green color. Darker colors indicate a larger absolute value of the correlation coefficient. The entire heatmap shows that the feature correlations are relatively low, and the dendrogram indicates that the associations occur on the features from different categories. Figure 4 provides the distributions of correlation values between features of the same feature type or image type. From Figure 4a, we can observe that the features within most subgroups have low association, while the only exception is the shape features. When sorting the features based on image types (Figure 4b), it becomes evident that certain features extracted from the same image type exhibit relatively higher correlations with each other. For example, the median correlation coefficient for features derived from gradient filter processed images is around 0.38, while this value is only 0.23 for the wavelet filter subgroup. Figure 5 plots the histogram of the absolute values of all the calculated correlation coefficients. It can be noted that 90.2 % of the correlation coefficient values are less than or equal to 0.5. The results indicate that the extracted 1373 raw features contain comprehensive information describing tumor attributes with negligible redundancy.

Table 2 summarizes the Linear-SVM based models' performance when only using one type of the features in modeling. The GLSZM features perform best among all different categories, yielding an AUC value of $0.638 \pm 0.094$ and an ACC value of 71.4 %. Meanwhile, the GLCM features achieve the second-highest performance, with AUC and ACC values of $0.633 \pm 0.094$ and 64.3 %, respectively. Similarly, the shape and first order features can still distinguish the positive cases from negative cases to a certain extent, resulting in AUC values of $0.617 \pm 0.095$ and $0.559 \pm 0.096$, respectively. In contrast, the remaining feature classes did not show significant discriminative powers. Figure 6 demonstrates the averaged weight of different feature types applying on the PCA-derived new features when all the 1373 features are used as the input of the Linear-SVM based model. It reflects that the features from different subcategories have approximately equal contributions, with the GLDM and GLRLM contributing slightly more than the others.

The ROC curves used to evaluate the performance of the models trained with 1373 features are plotted in Figure 7. The corresponding AUC value of the ROC curve for Linear-SVM model is $0.745 \pm 0.086$, which was produced by a ROC curve fitting program based on the maximum likelihood estimation method (ROCKIT, http://metz-roc.uchicago.edu/, University of Chicago). In contrast, the RBF-SVM based model produced a ROC curve that is generally higher than that of the Linear-SVM, enhancing its AUC value to 0.806±0.078. Figure 8 shows the confusion matrices for both models. As can be seen from Figure 8(a), out of the 10 cases predicted as "Suboptimal" by the Linear-SVM based model, 7 of them were truly sub-optimally debulked in IDS. This yields a positive predictive value (PPV) of 70% (7/10). On the other hand, among the 32 cases predicted to be "Optimal" by the Linear-SVM, 7 cases were confirmed receiving suboptimal debulking, corresponding to a negative prediction value (NPV) of 78.1% (25/32). Meanwhile, the RBF-SVM model (Figure 8b) offered PPV and NPV values of 81.8% (9/11) and 83.9% (26/31), respectively. Based on above calculations, the total accuracy (ACC) values of the linear and RBF kernel based SVM models reached to 76.2% (32/42) and 83.3% (35/42), respectively. This performance enhancement can be attributed to the presence of nonlinear prognostic information in the dataset, a characteristic that nonlinear RBF-SVM captures more efficiently than Linear-SVM.

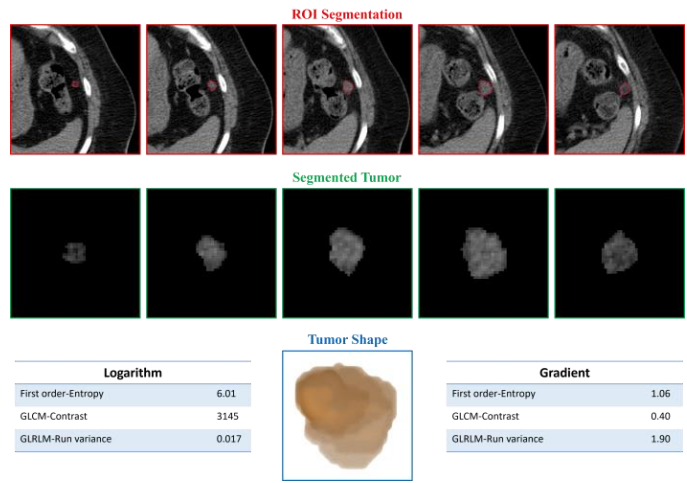

Fig. 2. One example of tumor segmentation and extracted features.

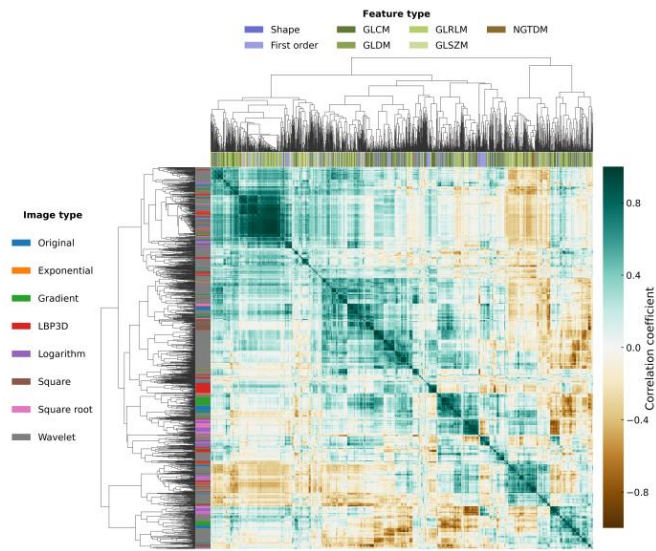

Fig. 3. The hierarchically clustered heatmap of Pearson's correlation coefficients between the extracted 1373 features.

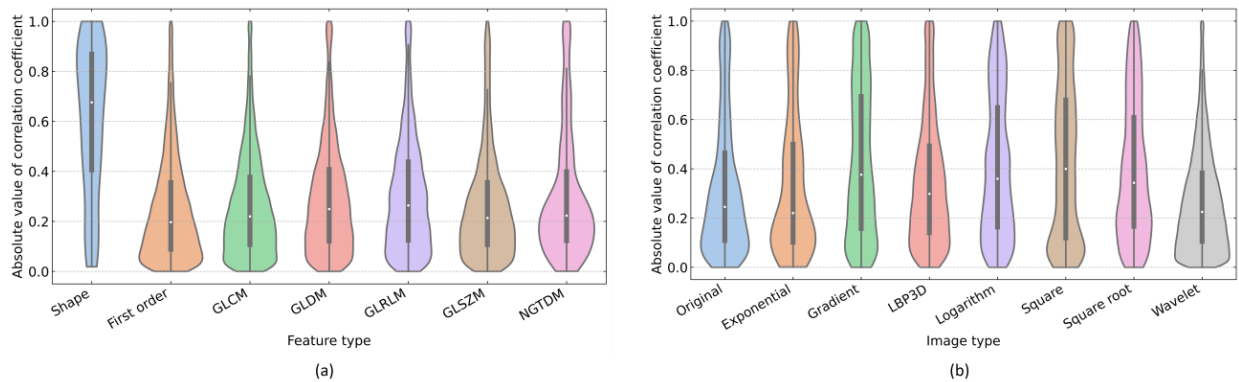

Fig. 4. The distribution of the absolute value of Pearson's correlation coefficients between features of the same feature type or image type.

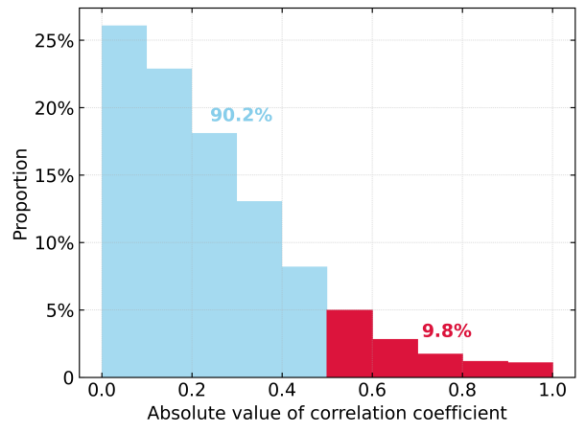

Fig.5. The histogram of the absolute value of correlation coefficients between the extracted 1373 features.

Table 2. The prediction performances of the Linear-SVM based models trained with different type of features.

|             | AUC               | ACC     |
|-------------|-------------------|---------|
| **Shape**   | 0.617 ± 0.095     | 64.3 %  |
| **First order** | 0.559 ± 0.096 | 66.7 %  |
| **GLCM**    | 0.633 ± 0.094     | 64.3 %  |
| **GLDM**    | 0.408 ± 0.092     | 52.4 %  |
| **GLRLM**   | 0.454 ± 0.094     | 59.5 %  |
| **GLSZM**   | **0.638 ± 0.094** | **71.4 %** |
| **NGTDM**   | 0.523 ± 0.096     | 38.1 %  |

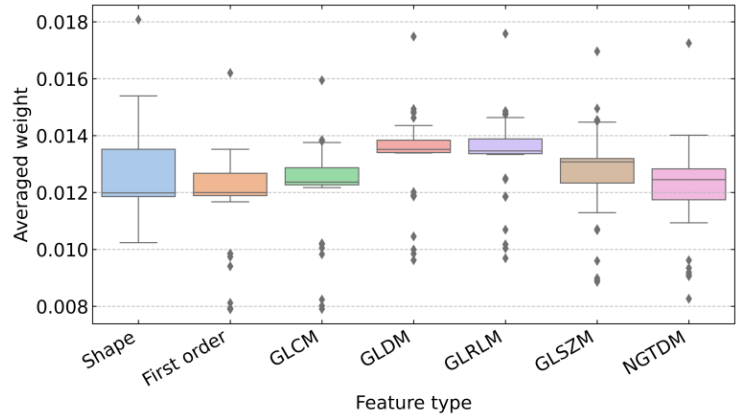

Fig. 6. Averaged weight of each feature type on the PCA-derived new features when training the Linear-SVM based model with the 1373 features.

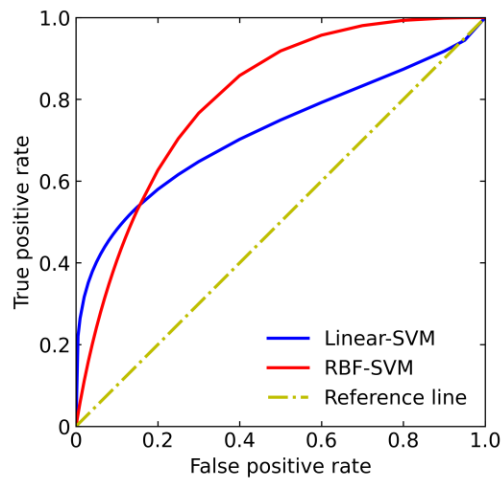

Fig. 7. The ROC curves for Linear-SVM and RBF-SVM based models trained with 1373 features, respectively.

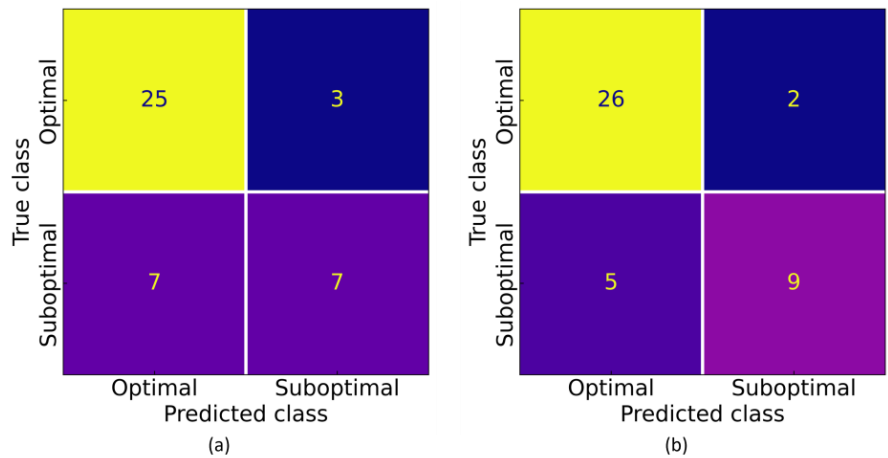

Fig. 8. The confusion matrices for (a) Linear-SVM and (b) RBF-SVM based models trained with 1373 features, respectively.

# 4. DISCUSSION

In this investigation, a novel image marker was initially developed and verified to categorize patients into optimal and suboptimal debulking groups prior to the NACT treatment. The major significance is this marker's potential to accurately stratify patients at an early stage. If subsequent research successfully validates this strategy, it could assist gynecologic oncologist to identify patients who may not benefit from the planned NACT treatment. Thus, alternative treatments can be chosen before chemotherapy is administered, allowing patients to avoid ineffective treatments and minimize the risk of severe side effects. Besides its significance, this study has following unique characteristics.

Different from the conventional image based NACT prognostic models, our proposed model utilized radiomics features to comprehensively quantify the tumor characteristics. The conventional models rely on either visual findings on the presence of malignancy sites (e.g. pleural effusion or mesentery deposits [14]) or some simple measurement depicted on the acquired images (e.g. tumor diameters [54, 55]). Despite the association between these features and cytoreduction, this method suffers from the inter- and intra- radiologist variability and time-varied diagnostic criteria [15, 56], which causes low robustness and unsatisfactory performance. In our study, the copious tumor information was thoroughly conveyed by the highly varied radiomics features. Followed by the PCA and SVM, this information was further identified and synthesized to generate a single clinical marker for outcome prediction. In a dataset containing 42 ovarian cancer patients, the marker achieved an overall accuracy of 76.2% and 83.3% when utilizing linear and RBF kernel based SVM algorithms, respectively. The results indicate the effectiveness of our proposed method. In addition, as compared to the other investigated molecular markers, our proposed method will not create additional financial burdens on the patients as CT imaging is a routine examination for ovarian cancer patients.

Furthermore, to the best of our knowledge, the computer feature pool in this study is by far one of the largest features pools for this specific or similar medical imaging tasks. We computed a total of 1373 radiomics features for NACT prognosis prediction, which can be divided into three main categories (e.g., shape, intensity, and texture). The prognostic capacities of various feature types were evaluated by training a Linear-SVM based model with each type individually. The result shows the five subcategories of texture features have varying levels of performance on this task: only GLSZM and GLCM indicate significant discriminative power on classifying optimal and suboptimal debulking in IDS (Table 2). Given that these two texture feature subgroups describe the connected or co-occurrence zones with the same grayscale level, it implies the NACT prognosis is more sensitive to such an intensity change. However, in the final feature cluster, all these features demonstrate an approximately equal contribution (Figure 6), which may be attributed by the fact that the features from different groups provide complementary information to the final marker. This observation also implies the necessity of radiomics method: each of these computed 1373 features characterizes the tumor heterogeneity uniquely, which is highly desirable in generating the final marker.

Despite the encouraging results, we acknowledge that the study has following limitations. First, the dataset only contains 42 patients from a single institution. The performance and robustness of our proposed algorithm should be further verified on a comprehensive multiple-institution database with diversified patients. Second, the model is developed based on the conventional SVM algorithm, and our study only used radiomics features. Some emerging technologies such as deep learning [57-59] were not adopted. One possible improvement of this study would be to combine radiomics and deep learning techniques together to further enhance the prediction accuracy [60]. Third, the feature repeatability and reproducibility were not investigated in current study. Utilizing radiomic features that remain consistent across various scanning setups, such as scanning equipment and acquisition software, can effectively decrease the risk of type I error and ensure a broader validity of the model [61]. Fourth, given the limited size of the dataset, we did not conduct model parameter uncertainty study. Despite these limitations, this investigation initially demonstrates the efficacy of using radiomic features for predicting NACT outcome in ovarian cancer treatment and lays a solid foundation for advancing precision treatment in future research.

## ACKNOWLEDGEMENTS

We acknowledge the support from the following funding agencies: National Institute of General Medical Sciences (NIGMS, Sub-project 8620, P20GM135009-02), Oklahoma Center for Advancement of Science and Technology (OCAST HR23-122).